\documentclass[letterpaper, 10 pt, conference]{ieeeconf}  
\IEEEoverridecommandlockouts                              

\makeatletter
\let\NAT@parse\undefined
\makeatother

\usepackage[numbers,sort]{natbib}
\usepackage{times}
\usepackage{graphicx}
\usepackage{bbm}
\usepackage{amsmath}
\usepackage{amssymb}
\usepackage{color}
\usepackage{tabularx}
\usepackage{hyperref}
\usepackage{algorithm}
\usepackage[noend]{algorithmic}

\title{\LARGE \bf
Closing the Sim-to-Real Loop: \\Adapting Simulation Randomization with Real World Experience}

\author{Yevgen Chebotar$^{1,2}$ \qquad Ankur Handa$^{1}$ \qquad Viktor Makoviychuk$^{1}$
\\[0.1em]
Miles Macklin$^{1,3}$ \qquad Jan Issac$^{1}$ \qquad Nathan Ratliff$^{1}$ \qquad Dieter Fox$^{1,4}$
\thanks{$^{1}$NVIDIA, USA
        }%
\thanks{$^{2}$University of Southern California, Los Angeles, CA, USA}%
\thanks{$^{3}$University of Copenhagen, Copenhagen, Denmark}
\thanks{$^{4}$University of Washington, Seattle, WA, USA} 
\thanks{$^@${\tt\small ychebota@usc.edu,\{ahanda,vmakoviychuk,
mmacklin,jissac,nratliff,dieterf\}@nvidia.com}} }

\begin{document}

\makeatletter
\let\@oldmaketitle\@maketitle
\renewcommand{\@maketitle}{\@oldmaketitle
  \includegraphics[width=\linewidth]
    {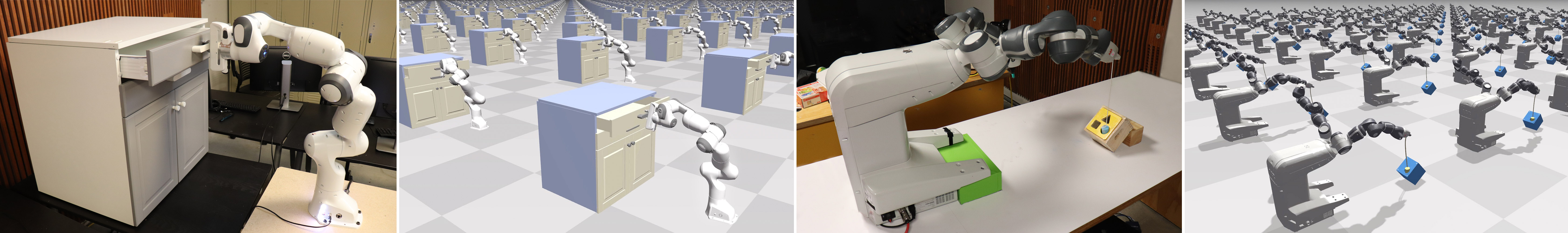} \\[0.35em]
  \refstepcounter{figure}\footnotesize{Fig.~\thefigure. Policies for opening a cabinet drawer and swing-peg-in-hole tasks trained by alternatively performing reinforcement learning with multiple agents in simulation and updating simulation parameter distribution using a few real world policy executions.}
  \label{fig:title} \medskip \vspace{-10pt}}
\makeatother

\maketitle
\thispagestyle{empty}
\pagestyle{empty}


\begin{abstract}
We consider the problem of transferring policies to the real world by training on a distribution of simulated scenarios. Rather than manually tuning the randomization of simulations, we adapt the simulation parameter distribution using a few real world roll-outs interleaved with policy training. In doing so, we are able to change the distribution of simulations to improve the policy transfer by matching the policy behavior in simulation and the real world.
We show that policies trained with our method are able to reliably transfer to different robots in two real world tasks: swing-peg-in-hole and opening a cabinet drawer. The video of our experiments can be found at \url{https://sites.google.com/view/simopt}.
\end{abstract}

\vspace{-1pt}
\section{Introduction}
\vspace{-2pt}
Learning continuous control in real world complex environments has seen a wide interest in the past few years and in particular focusing on learning policies in simulators and transferring them to the real world, as we still struggle with finding ways to acquire the necessary amount of experience and data in the real world directly.
While there have been recent attempts on learning by collecting large scale data directly on real robots~\cite{LevinePKIQ18, PintoG16, YahyaLKCL17, qtopt}, such an approach still remains challenging as collecting real world data is prohibitively laborious and expensive. Simulators offer several advantages, \textit{e.g.} they can run faster than real-time and allow for acquiring large diversity of  training data. However, due to the imprecise simulation models and lack of high fidelity replication of real world scenes, policies learned in simulations often cannot be directly applied on real world systems, a phenomenon also known as the \textit{reality gap}~\cite{Jakobi:etal:1995}. In this work, we focus on closing the reality gap by learning policies on distributions of simulated scenarios that are optimized for a better policy transfer.

Training policies on a large diversity of simulated scenarios by randomizing relevant parameters, also known as \textit{domain randomization}, has shown a considerable promise for the real world transfer in a range of recent works~\cite{TobinFRSZA17, Sadeghi:etal:RSS2017, James:etal:CoRL2017, handopenai}. 
However, design of the appropriate simulation parameter distributions remains a tedious task and often requires a substantial expert knowledge.
Moreover, there are no guarantees that the applied randomization would actually lead to a sensible real world policy as the design choices made in randomizing the parameters tend to be somewhat biased by the expertise of the practitioner. In this work, we apply a data-driven approach and use real world data to adapt simulation randomization such that the behavior of the policies trained in simulation better matches their behavior in the real world. Therefore, starting with some initial distribution of the simulation parameters, we can perform learning in simulation and use real world roll-outs of learned policies to gradually change the simulation randomization such that the learned policies  transfer better to the real world without requiring the exact replication of the real world scene in simulation. This approach falls into the domain of model-based reinforcement learning. 
However, we leverage recent developments in physics simulations to provide a strong prior of the world model in order to accelerate the learning process. Our system uses partial observations of the real world and only needs to compute rewards in simulation, therefore lifting the requirement for full state knowledge or reward instrumentation in the real world. 

\section{Related Work}
The problem of finding accurate models of the robot and the environment that can facilitate the design of robotic controllers in the real world dates back to the original works on system identification~\cite{Ljung, giri2010block}. In the context of reinforcement learning (RL), model-based RL explored optimizing policies using learned models~\cite{Deisenroth2013}. In~\cite{DeisenrothR11, DeisenrothRF11}, the data from real world policy executions is used to fit a probabilistic dynamics model, which is then used for learning an optimal policy. Although our work follows the general principle of model-based reinforcement learning, we aim at using a simulation engine as a form of parameterized model that can help us to embed prior knowledge about the world.  

Overcoming the discrepancy between simulated models and the real world has been addressed through identifying simulation parameters~\cite{KolevT15}, finding common feature representations of real and synthetic data~\cite{TzengDHFPLSD15}, using generative models to make synthetic images more realistic~\cite{graspgan}, fine-tuning the policies trained in simulation in the real world~\cite{RusuVRHPH17}, learning inverse dynamics models~\cite{ChristianoSMSBT16}, multi-objective optimization of task fitness and transferability to the real world~\cite{KoosMD10}, training on ensembles of dynamics models~\cite{MordatchLT15} and training on a large variety of simulated scenarios~\cite{TobinFRSZA17}. Domain randomization of textures was used in~\cite{Sadeghi:etal:RSS2017} to learn to fly a real quadcopter by training an image based policy entirely in simulation. Peng \textit{et al.}~\cite{PengetalICRA18} use randomization of physical parameters of the scene to learn a policy in simulation and transfer it to a real robot for pushing a puck to a target position. In~\cite{handopenai}, randomization of physical properties and object appearance is used to train a dexterous robotic hand to perform in-hand manipulation. Yu~\textit{et al.}~\cite{YuTLT17} propose to not only train a policy on a distribution of simulated parameters, but also learn a component that predicts the system parameters from the current states and actions, and use the prediction as an additional input to the policy. In \cite{MurTreGiePet18}, an  upper confidence bound on the estimated simulation optimization bias is used as a stopping criterion for a robust training with domain randomization.
In \cite{WulfmeierPA17}, an auxiliary reward is used to encourage policies trained in source and target environments to visit the same states.

Combination of system identification and dynamics randomization has been used in the past to learn locomotion for a real quadruped~\cite{Tan-RSS-18}, non-prehensile object manipulation~\cite{LowreyKDRT18} and in-hand object pivoting~\cite{AntonovaCSK17}.
In our work, we recognize domain randomization and system identification as powerful tools for training general policies in simulation. However, we address the problem of automatically learning simulation parameter distributions that improve policy transfer, as it remains challenging to do it manually. Furthermore, as also noticed in~\cite{lerrel}, simulators have an advantage of providing a full state of the system compared to partial observations of the real world, which is also used in our work for designing better reward functions.

The closest to our approach are the methods from~\cite{TanXBL16,ZhuKBB18,AAMAS13-Farchy,AAAI17-Hanna,RajeswaranGLR16} that propose to iteratively learn simulation parameters and train policies. In~\cite{TanXBL16}, an iterative system identification framework is used to optimize trajectories of a bipedal robot in simulation and calibrate the simulation parameters by minimizing the discrepancy between the real world and simulated execution of the trajectories. Although we also use the real world data to compute the discrepancy of the simulated executions, we are able to use partial observations of the real world instead of the full states and we concentrate on learning general policies by finding simulation parameter distribution that leads to a better transfer without the need for exact replication of the real world environment. \cite{ZhuKBB18}~suggests to optimize the simulation parameters such that the value function is well approximated in simulation without replicating the real world dynamics. We also recognize that exact replication of the real world dynamics might not be feasible, however a suitable randomization of the simulated scenarios can still lead to a successful policy transfer. In addition, our approach does not require estimating the reward in the real world, which might be challenging if some of the reward components can not be observed. \cite{AAMAS13-Farchy} and \cite{AAAI17-Hanna} consider grounding the simulator using real world data. However, \cite{AAMAS13-Farchy} requires a human in the loop to select the best simulation parameters, and~\cite{AAAI17-Hanna} needs to fit additional models for the real robot forward dynamics and simulator inverse dynamics. 
Finally, our work is closest to the adaptive EPOpt framework of Rajeswaran \textit{et al.}~\cite{RajeswaranGLR16}, which optimizes a policy over an ensemble of models and adapts the model distribution using data from the target domain. EPOpt optimizes a risk-sensitive objective to obtain robust policies, whereas we optimize the average performance which is a risk-neutral objective. Additionally, EPOpt updates the model distribution by employing Bayesian inference with a particle filter, whereas we update the model distribution using an iterative KL-divergence constrained procedure. More importantly, they focus on simulated environments while in our work, we develop an approach that is shown to work in the real world and apply it to two real robot tasks.

\section{Closing the Sim-to-Real Loop}
\label{sec:method}
\subsection{Simulation randomization}
Let $\mathcal{M} = (S, A, P, R, p_0, \gamma, T)$ be a finite-horizon Markov Decision Process (MDP), where $S$ and $A$ are state and action spaces, $P: S \times A \times S \rightarrow \mathbb{R}_{+}$ is a state-transition probability function or probabilistic system dynamics, $R: S \times A \rightarrow \mathbb{R}$ a reward function, $p_0: S \rightarrow \mathbb{R}_{+}$ an initial state distribution, $\gamma$ a reward discount factor, and $T$ a fixed horizon. Let $\tau = (s_0, a_0, \dots, s_T, a_T)$ be a trajectory of states and actions and $R(\tau) = \sum_{t=0}^T \gamma^t R(s_t,a_t)$ the trajectory reward. The goal of reinforcement learning methods is to find parameters $\theta$ of a policy $\pi_\theta(a | s)$ that maximize the expected discounted reward over trajectories induced by the policy: 
$\mathbb{E}_{\pi_\theta}[R(\tau)]$ where $ s_0\sim p_0, s_{t+1}\sim P(s_{t+1} | s_t, a_t)$ and $a_t\sim \pi_\theta(a_t | s_t)$.

In our work, the system dynamics are either induced by a simulation engine or real world. As the simulation engine itself is deterministic, a reparameterization trick~\cite{KingmaW13} can be applied to introduce probabilistic dynamics. In particular, we define a distribution of simulation parameters $\xi \sim p_\phi(\xi)$ parameterized by $\phi$. The resulting probabilistic system dynamics of the simulation engine are $P_{\xi\sim p_\phi} = P(s_{t+1} | s_t, a_t, \xi)$.

As it was shown in \cite{TobinFRSZA17, Sadeghi:etal:RSS2017, handopenai}, it is possible to design a distribution of simulation parameters $p_\phi(\xi)$, such that a policy trained on $P_{\xi\sim p_\phi}$ would perform well on a real world dynamics distribution. This approach is also known as \textit{domain randomization} and the policy training maximizes the expected reward under the dynamics induced by the distribution of simulation parameters $p_\phi(\xi)$:
\begin{equation}
\label{eq:domain_random}
\max_\theta \mathbb{E}_{P_{\xi\sim p_\phi}} \left[\mathbb{E}_{\pi_\theta}[R(\tau)]\right]
\end{equation}
Domain randomization requires a significant expertise and tedious manual fine-tuning to design the simulation parameter distribution $p_\phi(\xi)$. Furthermore, as we show in our experiments, it is often disadvantageous to use overly wide distributions of simulation parameters as they can include scenarios with infeasible solutions that hinder successful policy learning, or lead to exceedingly conservative policies. Instead, in the next section, we present a way to automate the learning of $p_\phi(\xi)$ that makes it possible to shape a suitable randomization without the need to train on very wide distributions.

\subsection{Learning simulation randomization}
\begin{figure}[t]
\vspace{12pt}
\begin{center}
\centerline{\includegraphics[width=0.99\columnwidth]{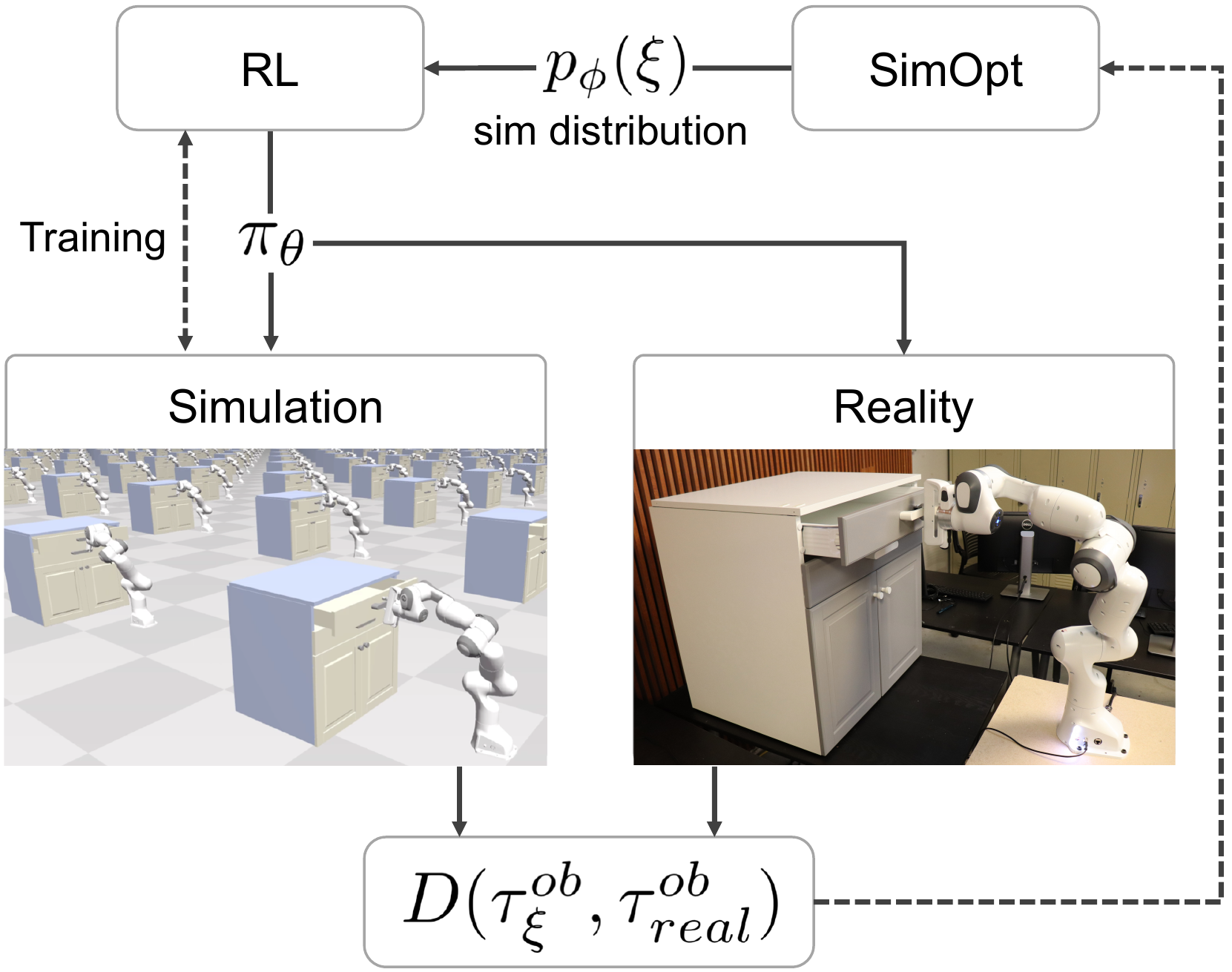}}
\vspace{-2pt}
\caption{The pipeline for optimizing the simulation parameter distribution. After training a policy on current distribution, we sample the policy both in the real world and for a range of parameters in simulation. The discrepancy between the simulated and real observations is used to update the simulation parameter distribution in SimOpt.}
\label{fig:simopt_graph}
\end{center}
\vspace{-15pt}
\end{figure} 

The goal of our framework is to find a distribution of simulation parameters that brings observations or partial observations induced by the policy trained under this distribution closer to the observations of the real world. Let $\pi_{\theta,p_\phi}$ be a policy trained under the simulated dynamics distribution $P_{\xi\sim p_\phi}$ as in the objective (\ref{eq:domain_random}), and let
$D(\tau^{ob}_{\xi}, \tau^{ob}_{real})$ be a measure of discrepancy between real world observation trajectories $\tau^{ob}_{real}=(o_{0, real} \dots, o_{T, real})$ and simulated observation trajectories $\tau^{ob}_{\xi}=(o_{0, \xi} \dots, o_{T, \xi})$ sampled using policy $\pi_{\theta,p_\phi}$ and the dynamics distribution $P_{\xi\sim p_\phi}$. It should be noted that the inputs of the policy $\pi_{\theta,p_\phi}$ and observations used to compute $D(\tau^{ob}_{\xi}, \tau^{ob}_{real})$ are not required to be the same. The goal of optimizing the simulation parameter distribution is to minimize the following objective:
\vspace{1pt}
\begin{equation}
\label{eq:simopt}
\min_{\phi}   \mathbb{E}_{P_{\xi\sim p_\phi}} \left[\mathbb{E}_{\pi_{\theta,p_\phi}}\left[D(\tau^{ob}_\xi, \tau^{ob}_{real})\right]\right]
\end{equation}
\vspace{1pt}This optimization would entail training and real robot evaluation of the policy $\pi_{\theta,p_\phi}$ for each  $\phi$. This would require a large amount of RL iterations and more critically real robot trials. Hence, we develop an iterative approach to approximate the optimization by training a policy $\pi_{\theta, p_{\phi_i}}$ on the simulation parameter distribution from the previous iteration and using it for both, sampling the real world observations and optimizing the new simulation parameter distribution $p_{\phi_{i+1}}$:
\begin{align}
\label{eq:simopt_iter}
\min_{\phi_{i+1}}\, & \mathbb{E}_{P_{\xi_{i+1}\sim p_{\phi_{i+1}}}}  \left[\mathbb{E}_{\pi_{\theta,p_{\phi_i}}} \left[D(\tau^{ob}_{\xi_{i+1}}, \tau^{ob}_{real})\right]\right]\\
&\mbox{s.t.} \quad D_{KL}\left(p_{\phi_{i+1}} \| p_{\phi_{i}}\right) \leq \epsilon, \nonumber
\end{align}
where we introduce a KL-divergence step $\epsilon$ between the old simulation parameter distribution $p_{\phi_i}$ and the updated distribution $p_{\phi_{i+1}}$ to avoid going out of the trust region of the policy $\pi_{\theta, p_{\phi_i}}$  trained on the old simulation parameter distribution. Fig.~\ref{fig:simopt_graph} shows the general structure of our algorithm that we call \textit{SimOpt}.

\begin{figure}[t]
\begin{algorithm}[H]
\begin{algorithmic}[1]
\STATE{$p_{\phi_0} \gets \text{Initial simulation parameter distribution}$}
\STATE{$\epsilon \gets \text{KL-divergence step for updating } p_{\phi}$}
\FOR{iteration $i \in \{0,\dots,N\}$}
 \STATE{$\textsf{env} \gets \textsf{Simulation}(p_{\phi_i})$}
 \STATE {$\pi_{\theta,p_{\phi_i}} \gets \textsf{RL}(\textsf{env})$}
 \STATE{$\tau^{ob}_{real} \sim \textsf{RealRollout}(\pi_{\theta,p_{\phi_i}})$} 
 \STATE{$\xi \sim \textsf{Sample}(p_{\phi_i})$}
 \STATE{$\tau^{ob}_\xi \sim \textsf{SimRollout}(\pi_{\theta,p_{\phi_i}}, \xi)$}
 \STATE{$c(\xi) \gets D(\tau^{ob}_\xi, \tau^{ob}_{real})$}
 \STATE{$p_{\phi_{i+1}} \gets \textsf{UpdateDistribution}(p_{\phi_i}, \xi,c(\xi),\epsilon)$}
\ENDFOR
\end{algorithmic}
\caption{SimOpt framework}
\label{algo:simopt}
\end{algorithm}
\vspace{-23pt}
\end{figure}

\subsection{Implementation}
Here we describe particular implementation choices for the components of our framework used in this work. However, it should be noted that each of the components is replaceable. Algorithm~\ref{algo:simopt} describes the order of running all the components in our implementation.   
The RL training is performed on a GPU based simulator using a parallelized version of proximal policy optimization (PPO)~\cite{SchulmanWDRK17} on a multi-GPU cluster~\cite{Liang:etal:CoRL2018}. We parameterize our simulation parameter distribution as a Gaussian, \textit{i.e.} $p_\phi(\xi) \sim \mathcal{N}(\mu, \Sigma)$ with $\phi = (\mu, \Sigma)$. We choose weighted $\ell_1$ and $\ell_2$ norms between simulation and real world observations for our observation discrepancy function $D$:
\begin{align}
 &D(\tau^{ob}_\xi, \tau^{ob}_{real}) = \\ \nonumber
 &w_{\ell_1} \sum_{i=0}^T |W (o_{i, \xi} - o_{i, real})| +  w_{\ell_2} \sum_{i=0}^T \|W (o_{i, \xi} - o_{i, real})\|^2_2, \nonumber
\end{align}
where $w_{\ell_1}$ and $w_{\ell_2}$ are the weights of the $\ell_1$ and $\ell_2$ norms, and $W$ are the importance weights  for each observation dimension. 
We additionally apply a Gaussian filter to the distance computation to account for misalignments of the trajectories. 
 
As we use a non-differentiable simulator we employ a sampling-based gradient-free algorithm based on relative entropy policy search~\cite{PetersMA10} for optimizing the objective (\ref{eq:simopt_iter}), which is able to perform updates of $p_\phi$ with an upper bound on the KL-divergence step. By doing so, the simulator can be treated as a black-box, as in this case $p_\phi$ can be optimized directly by only using samples $\xi \sim p_\phi$ and the corresponding costs $c(\xi)$ coming from $D(\tau^{ob}_{\xi}, \tau^{ob}_{real})$. Sampling of simulation parameters and the corresponding policy roll-outs is highly parallelizable, which we use in our experiments to evaluate large amounts of simulation parameter samples.

As noted above, single components of our framework can be exchanged. In case of availability of a differentiable simulator, the objective (\ref{eq:simopt_iter}) can be defined as a loss function for optimizing with gradient descent. Furthermore, for cases where $\ell_1$ and $\ell_2$ norms are not applicable, we can employ other forms of discrepancy functions, e.g. to account for potential domain shifts between observations~\cite{TzengDHFPLSD15,TzengHDS15,tcn}. Alternatively, real world and simulation data can be additionally used to train $D(\tau^{ob}_\xi, \tau^{ob}_{real})$ to discriminate between the observations by minimizing the prediction loss of classifying observations as simulated or real, similar to the discriminator training in the generative adversarial framework~\cite{GoodfellowPMXWOCB14,HoE16,HausmanCSSL17}.  Finally, a higher-dimensional generative model $p_\phi(\xi)$ can be employed to provide a multi-modal randomization of the simulated environments.

\vspace{5pt}
\section{Experiments}
 \vspace{2pt}
In our experiments we aim at answering the following questions: (1)~How does our method compare to standard domain randomization? (2)~How learning a simulation parameter distribution compares to training on a very wide  parameter distribution? (3)~How many \textit{SimOpt} iterations and real world trials are required for a successful transfer of robotic manipulation policies? (4)~Does our method work for different real world tasks and robots? 

We start by performing an ablation study in simulation by transferring policies between scenes with different initial state distributions, such as different poses of the cabinet in the drawer opening task. We demonstrate that updating the distribution of simulation parameters leads to a successful policy transfer in contrast to just using an initial distribution of the parameters without any updates as done in standard domain randomization. As we observe, training on very wide parameter distributions is significantly more difficult and prone to fail compared to initializing with a conservative parameter distribution and updating it using \textit{SimOpt} afterwards.

Next, we show that we can successfully transfer policies to real robots, such as ABB Yumi and Franka Panda, for complex articulated tasks such as cabinet drawer opening, and tasks with non-rigid bodies and complex dynamics, such as swing-peg-in-hole task with the peg swinging on a soft rope. The policies can be transferred with a very small amount of real robot trials and leveraging large-scale training on a multi-GPU cluster.

 \vspace{3pt}
\subsection{Tasks}
 \vspace{1pt}
We evaluate our approach on two robot manipulation tasks: cabinet drawer opening and swing-peg-in-hole. 
\subsubsection{Swing-peg-in-hole}
The goal of this task is to put a peg attached to a robot hand on a rope into a hole placed at a 45 degrees angle. Manipulating a soft rope leads to a swinging motion of the peg, which makes the dynamics of the task more challenging. The task set up in the simulation and real world using a 7-DoF Yumi robot from ABB is depicted in Fig.~\ref{fig:title} on the right. Our observation space consists of 7-DoF arm joint configurations and 3D position of the peg. The \textit{reward function} for the RL training in simulation includes the distance of the peg from the hole, angle alignment with the hole and a binary reward for solving the task.

\subsubsection{Drawer opening}
In the drawer opening task, the robot has to open a drawer of a cabinet by grasping and pulling it with its fingers. This task involves an ability to handle contact dynamics when grasping the drawer handle. For this task, we use a 7-DoF Panda arm from Franka Emika. Simulated and real world settings are shown in Fig.~\ref{fig:title} on the left. This task is operated on a 10D observation space: 7D robot joint angles and 3D position of the cabinet drawer handle. The \textit{reward function} consists of the distance penalty between the handle and end-effector positions, the angle alignment of the end-effector and the drawer handle, the opening distance of the drawer and an indicator function ensuring that both robot fingers are on the handle.

We would like to emphasize that our method does not require the full state information of the real world, \textit{e.g.} we do not need to estimate the rope diameter, rope compliance \textit{etc.} to update the simulation parameter distribution in the swing-peg-in-hole task. The output of our policies consists of 7 joint velocity commands and an additional gripper command for the drawer opening task.

\begin{figure}[t]
\begin{center}
\vspace{10pt}
\centerline{\includegraphics[width=\columnwidth,trim=0 0 0 0.5cm,clip]{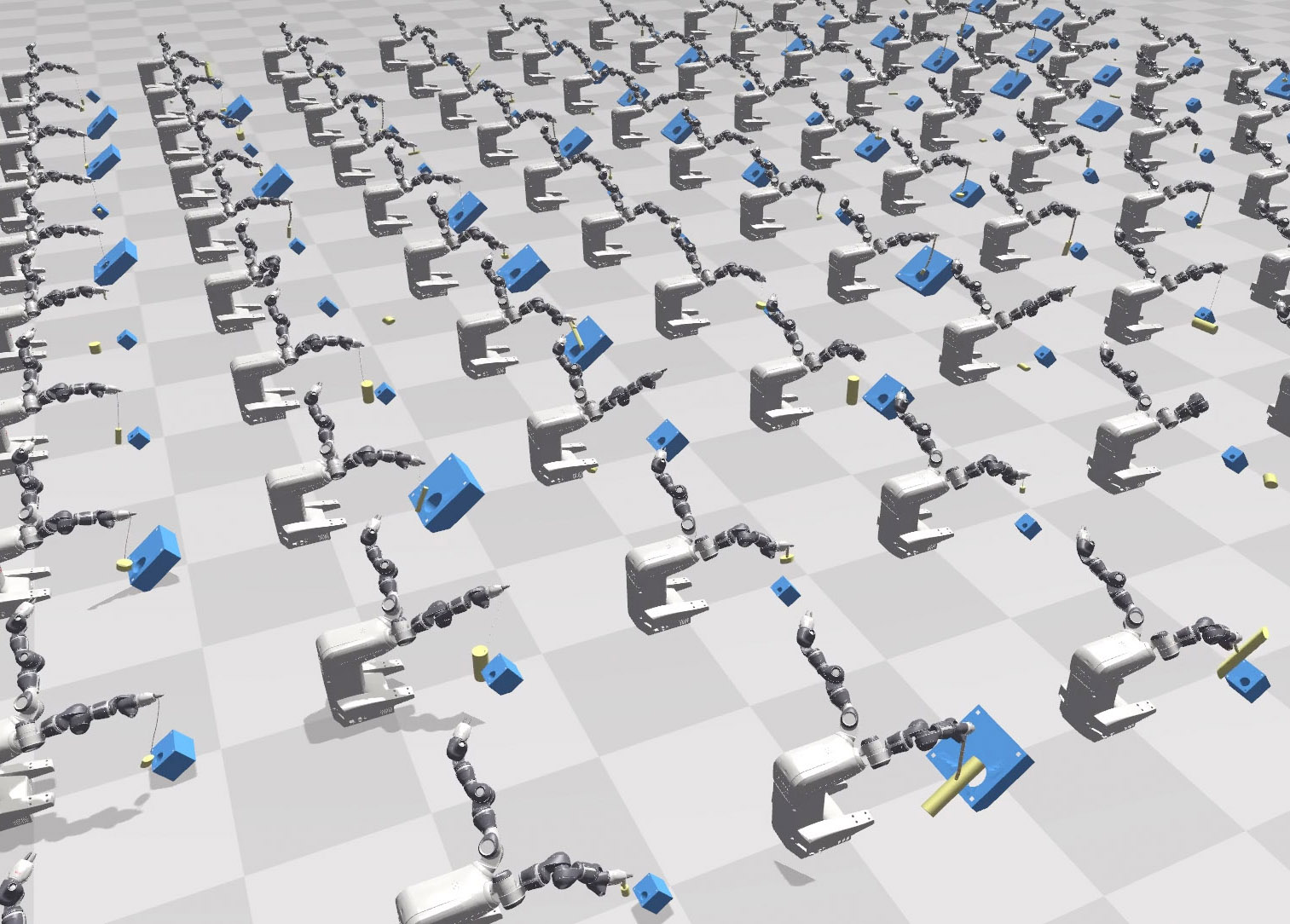}}
\caption{An example of a wide distribution of simulation parameters in the swing-peg-in-hole task where it is not possible to find a solution for many of the task instances.}
\label{fig:peghole_widedist}
\end{center}
\vspace{-10pt}
\end{figure} 

\vspace{3pt}
\subsection{Simulation engine}
\vspace{2pt}
We use NVIDIA Flex as a high-fidelity GPU based physics simulator that uses maximal coordinate representation to simulate rigid body dynamics. Flex allows a highly parallel implementation and can simulate multiple instances of the scene on a single GPU. We use the multi-GPU based RL infrastructure developed in~\cite{Liang:etal:CoRL2018} to leverage the highly parallel nature of the simulator.

\subsection{Comparison to standard domain randomization}
\vspace{-1pt}
We aim at understanding what effect a wide simulation parameter distribution can have on learning robust policies, and how we can improve the learning performance and the transferability of the policies using our method to adjust simulation randomization. Fig.~\ref{fig:peghole_widedist} shows an example of training a policy on a significantly wide distribution of simulation parameters for the swing-peg-in-hole task. In this case, peg size, rope properties and size of the peg box were randomized.
As we can observe, a large part of the randomized instances does not have a feasible solution, \textit{i.e.} when the peg is too large for the hole or the rope is too short. Finding a suitably wide parameter distribution would require  manual fine-tuning of the randomization parameters.

\begin{figure}[t]
\begin{center}
\vspace{7pt}
\centerline{\includegraphics[width=\columnwidth,trim=1cm 0.4cm 2.6cm 1.5cm,clip]{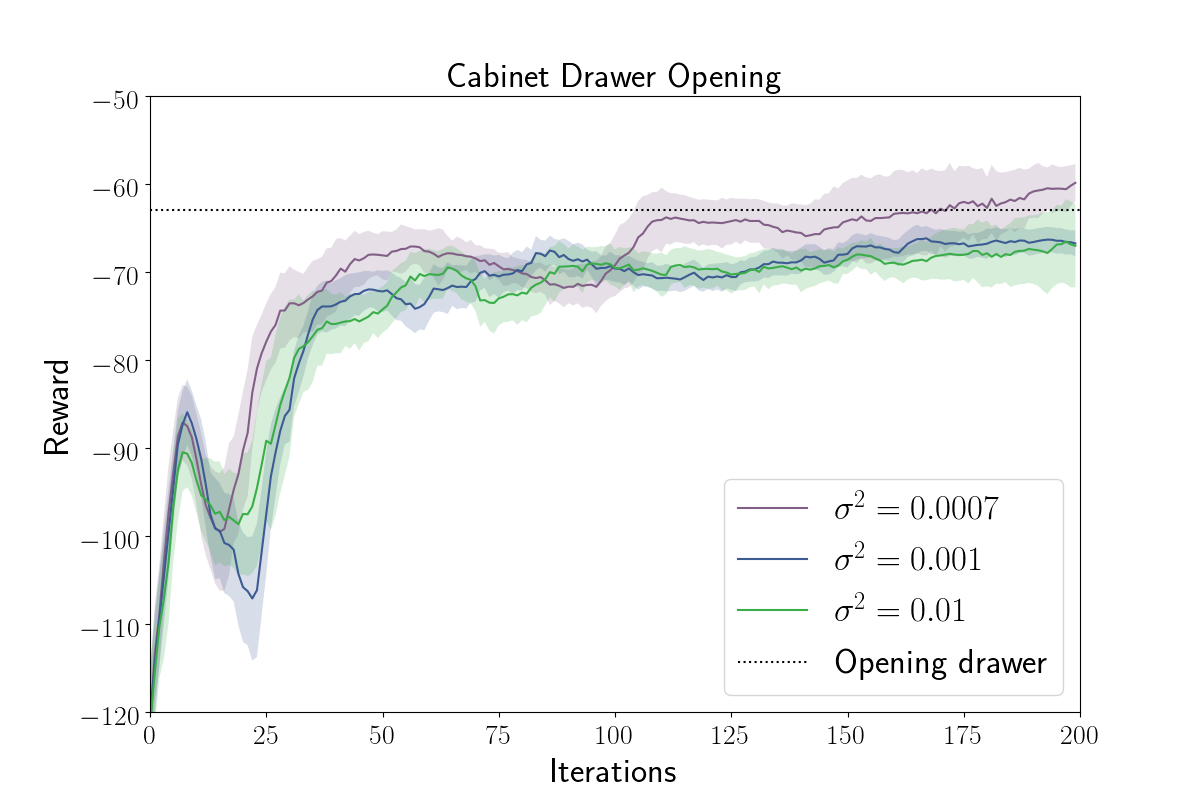}}
\vspace{-4pt}
\caption{Performance of the policy training with standard domain randomization for different variances of the distribution of the cabinet position along the X-axis in the drawer opening task.}
\label{fig:franka_cabinet}
\end{center}
\vspace{-7pt}
\end{figure}

Moreover, learning performance of standard domain randomization depends strongly on the variance of the parameter distribution. We investigate this in a simulated cabinet drawer opening task with a Franka arm which is placed in front of a cabinet. We randomize the position of the cabinet along the lateral direction (X-coordinate) while keeping all other simulation parameters constant. We train our policies on a 2 layer neural network with fully connected layers of 64 units each with PPO for 200 iterations. As we increase the variance of the cabinet position, we observe that the policies learned tend to be conservative \textit{i.e.} they do end up reaching the handle of the drawer but fail to open it. This is shown in Fig.~\ref{fig:franka_cabinet} where we plot the reward as a function of number of iterations used to train the RL policy. We start with a standard deviation of 2cm ($\sigma^2 = 7e-4$) and increase it to 10cm ($\sigma^2 = 0.01$). As shown in the plot, the policy is sensitive to the choice of this parameter and only manages to open the drawer when the standard deviation is 2cm. We note that the reward difference may not seem that significant but realize that it is dominated by the reaching reward. Increasing variance further, in an attempt to cover a wider operating range, can often lead to simulating unrealistic scenarios and catastrophic breakdown of the physics simulation with various joints of the robot reaching their limits. We also observed that the policy is extremely sensitive to the variance in all three axes of the cabinet position \textit{i.e.} policy only ever converges when the standard deviation is 2cm and fails to learn even reaching the handle otherwise.

\begin{figure}[t]
\begin{center}
\vspace{-4pt}
\centerline{\includegraphics[width=\columnwidth,trim=28pt 0 70pt 25pt,clip]{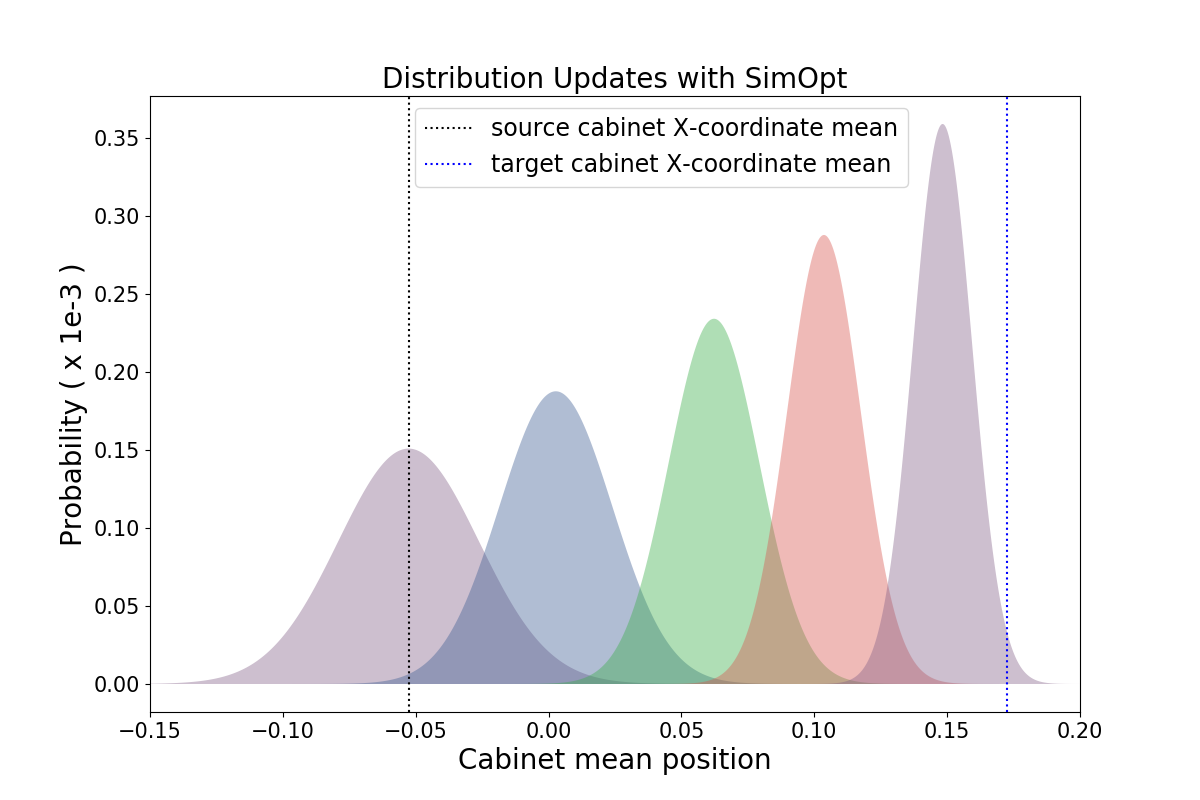}}
\vspace{-7pt}
\caption{Initial distribution of the cabinet position in the source environment, located at extreme left, slowly starts to change to the target environment distribution as a function of running 5 iterations of SimOpt.}
\label{fig:franka_cabinet_sim2sim_reps_updates}
\end{center}
\vspace{-19pt}
\end{figure}

\begin{figure*}[t]
\vspace{5pt}
\centerline{\includegraphics[width=\textwidth]{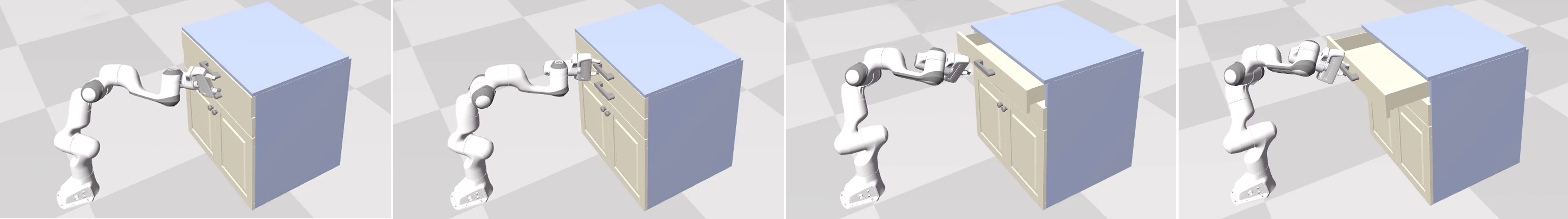}}
\vspace{-5pt}
\caption{Policy performance in the target drawer opening environment trained on randomized simulation parameters at different iterations of SimOpt. As the source environment distribution gets adjusted, the policy transfer improves until the robot can successfully solve the task in the fourth SimOpt iteration.}
\label{fig:franka_cabinet_sim2sim}
\vspace{-3pt}
\end{figure*}

\begin{figure*}[t]
\begin{center}
\centerline{\includegraphics[width=\textwidth]{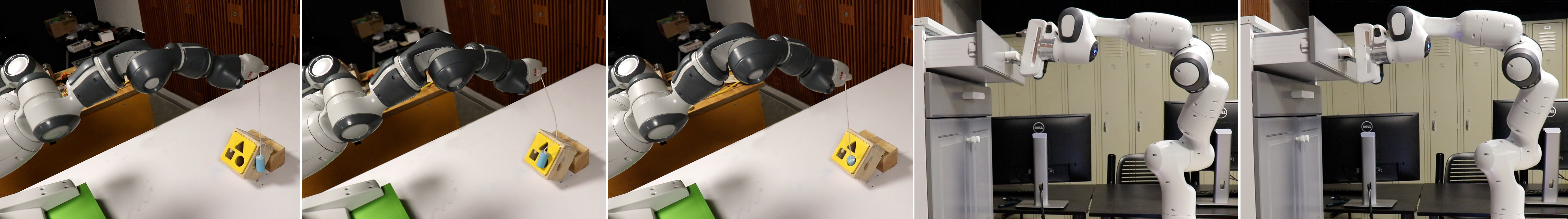}}
\vspace{-5pt}
\caption{Running policies trained in simulation at different iterations of SimOpt for real world swing-peg-in-hole and drawer opening tasks. \textit{Left:} SimOpt adjusts physical parameter distribution of the soft rope, peg and the robot, which results in a successful execution of the task on a real robot after two SimOpt iterations. \textit{Right:} SimOpt adjusts physical parameter distribution of the robot and the drawer. Before updating the parameters, the robot pushes too much on the drawer handle with one of its fingers, which leads to opening the gripper. After one SimOpt iteration, the robot can better control its gripper orientation, which leads to an accurate task execution.}
\label{fig:yumi_franka_screens}
\end{center}
\vspace{-19pt}
\end{figure*} 

In our next set of experiments, we show that our method is able to perform policy transfer from the source to target drawer opening scene where position of the cabinet in the target scene is offset by a distance of 15cm and 22cm. Such large distances would have required the standard deviation of the cabinet position to be at least 10cm for any na\"ive domain randomization based training which fails to produce a policy that opens the drawer as shown in Fig.~\ref{fig:franka_cabinet}. 
The policy is first trained with RL on a conservative initial simulation parameter distribution. Afterwards, it is run on the target scene to collect roll-outs. These roll-outs are then used to perform several \textit{SimOpt} iterations to optimize simulation parameters that best explain the current roll-outs.
We noticed that the RL training can be sped up by initializing the policy with the weights from the previous \textit{SimOpt} iteration, effectively reducing the number of needed PPO iterations from 200 to 10 after the first \textit{SimOpt} iteration. The whole process is repeated until the learned policy starts to successfully open the drawer in the target scene. We found that it took overall 3 iterations of doing RL and \textit{SimOpt} to learn to open the drawer when the cabinet was offset by 15cm. 
We further note that the number of iterations increases to 5 as we increase the target cabinet distance to 22cm  highlighting that our method is able to operate on a wider range of mismatch between the current scene and the target scene. Fig.~\ref{fig:franka_cabinet_sim2sim_reps_updates} shows how the source distribution variance adapts to the target distribution variance for this experiment and Fig.~\ref{fig:franka_cabinet_sim2sim} shows that our method starts with a conservative guess of the initial distribution of the parameters and changes it using target scene roll-outs until policy behavior in target and source scenes starts to match.

\subsection{Real robot experiments}
In our real robot experiments, \textit{SimOpt} is used to learn simulation parameter distribution of the manipulated objects and the robot. We run our experiments on 7-DoF Franka Panda and ABB Yumi robots. The RL training and \textit{SimOpt} simulation parameter sampling is performed using a cluster of 64 GPUs for running the simulator with 150 simulated agents per GPU. In the real world, we use object tracking with DART~\cite{Schmidt:etal:RSS2014} to continuously track the 3D positions of the peg in the swing-peg-in-hole task and the handle of the cabinet drawer in the drawer opening task, as well as initialize positions of the peg box and the cabinet in simulation. DART operates on depth images and requires 3D articulated models of the objects. We learn multi-variate Gaussian distributions of the simulation parameters parameterized by a mean and a full covariance matrix, and perform several updates of the simulation parameter distribution per \textit{SimOpt} iteration using the same real world roll-outs to minimize the number of real world trials.

\subsubsection{Swing-peg-in-hole}
Fig.~\ref{fig:yumi_franka_screens}~(left) demonstrates the behavior of real robot execution of the policy trained in simulation over 3 iterations of \textit{SimOpt}. At each iteration, we perform 100 iterations of RL in approximately 7 minutes and 3 roll-outs on the real robot using the currently trained policy to collect real world observations. Then, we run 3 update steps of the simulation parameter distribution with 9600 simulation samples per update. In the beginning, the robot misses the hole due to the discrepancy of the simulation parameters and the real world. After a single \textit{SimOpt} iteration, the robot is able to get much closer to the hole, however not being able to insert the peg as it requires a slight angle to go into the hole, which is non-trivial to achieve using a soft rope. Finally, after two \textit{SimOpt} iterations, the policy trained on a resulting simulation parameter distribution is able to swing the peg into the hole in $90\%$ of the times when evaluated on 20 trials. 

We observe that the most significant changes of the simulation parameter distribution occur in the physical parameters of the rope that influence its dynamical behavior and the robot parameters that influence the policy behavior, such as scaling of the policy actions. More details on the initial and updated Gaussian  distribution parameters can be found in Appendix \ref{app:sim_params}. 
Fig.~\ref{fig:yumi_cov_heatmap} shows the development of the covariance matrix over the iterations. We can observe some correlation in the top left block of the matrix, which corresponds to the robot joint compliance and damping values. This reflects the fact that these values have somewhat opposite effect on the robot behavior, i.e. if we overshoot in the compliance we can compensate with increased damping. 

\begin{figure}[t]
\begin{center}
\vspace{4pt}
\includegraphics[width=0.245\columnwidth,trim=0 0 0 0,clip]{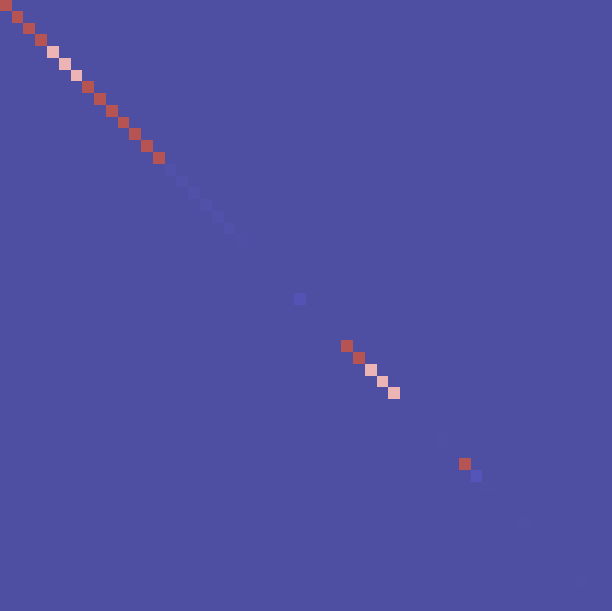}\hspace{-2.6pt}
\includegraphics[width=0.245\columnwidth,trim=0 0 0 0,clip]{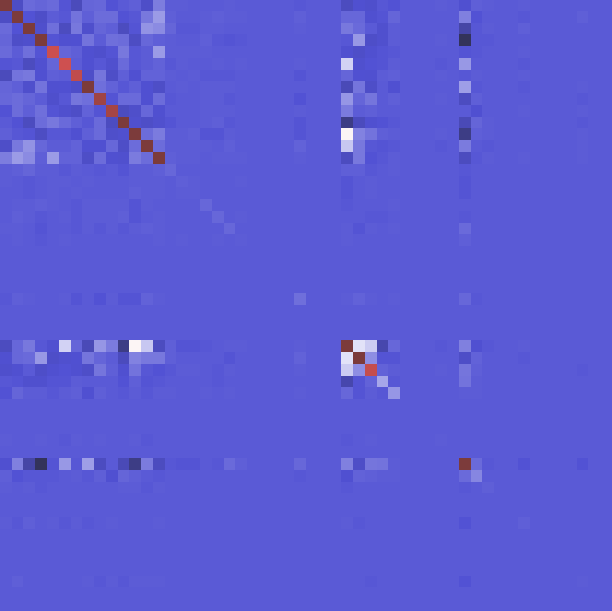}\hspace{-2.5pt}
\includegraphics[width=0.245\columnwidth,trim=0 0 0 0,clip]{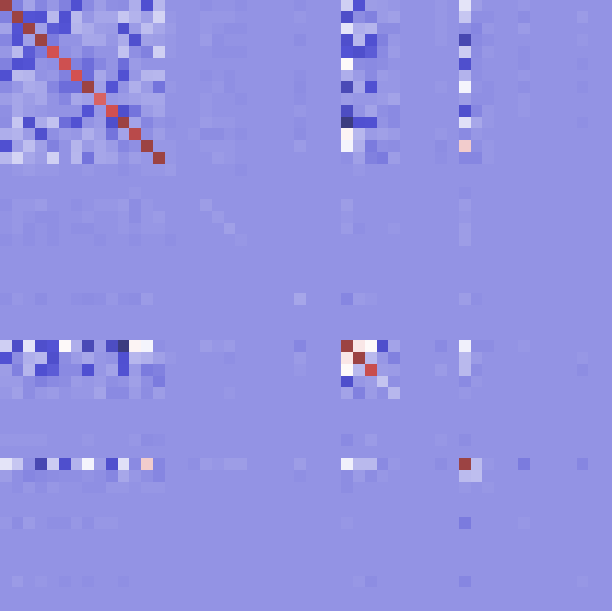}\hspace{-2.5pt}
\includegraphics[width=0.245\columnwidth,trim=0 0 0 0,clip]{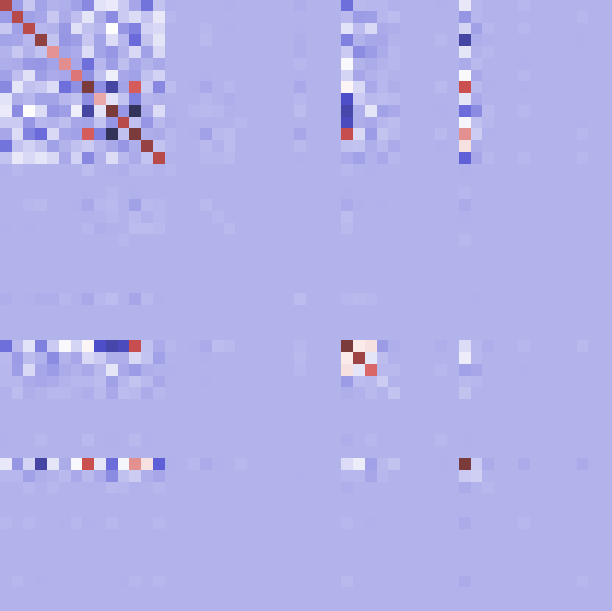}\hspace{-2.4pt}
\vspace{-4pt}
\caption{Covariance matrix heat maps over 3 SimOpt updates of the swing-peg-in-hole task beginning with the initial covariance matrix.}
\label{fig:yumi_cov_heatmap}
\end{center}
\vspace{-16pt}
\end{figure}

\subsubsection{Drawer opening}
For drawer opening, we learn a Gaussian distribution of the robot and cabinet simulation parameters. More details on the learned distribution and its initialization are provided in Appendix \ref{app:sim_params}. 
Fig.~\ref{fig:yumi_franka_screens}~(right) shows the drawer opening behavior before and after performing a \textit{SimOpt} update. During each \textit{SimOpt} iteration, we run 200 iterations of RL for approximately 22 minutes, perform 3 real robot roll-outs and run 20 update steps of the simulation distribution using 9600 samples per update step. Before updating the parameter distribution, the robot is able to reach the handle and start opening the drawer. However, it cannot exactly replicate the learned behavior from simulation and does not keep the gripper orthogonal to the drawer, which results in pushing too much on the handle from the bottom with one of the robot fingers. As the finger gripping force is limited, the fingers begin to open due to a larger pushing force. After adjusting the simulation parameter distribution  that includes robot and drawer properties, the robot is able to better control its gripper orientation and by evaluating on 20 trials can open the drawer at all times keeping the gripper orthogonal to the handle.

\section{Conclusions}
Closing the simulation to reality transfer loop is an important component for a robust transfer of robotic policies. In this work, we demonstrated that adapting simulation randomization using real world data can help in learning simulation parameter distributions that are particularly suited for a successful policy transfer without the need for exact replication of the real world environment. In contrast to trying to learn policies using very wide distributions of simulation parameters, which can simulate infeasible scenarios, we are able to start with distributions that can be efficiently learned with reinforcement learning, and modify them for a better transfer to the real world scenario. Our framework does not require full state of the real environment and reward functions are only needed in simulation. We showed that updating simulation distributions is possible using partial observations of the real world while the full state still can be used for the reward computation in simulation. We evaluated our approach on two real world robotic tasks and showed that policies can be transferred with only a few iterations of simulation updates using a small number of real robot trials.

In this work, we applied our method to learning uni-modal simulation parameter distributions. We plan to extend our framework to multi-modal distributions and more complex generative simulation models in future work. Furthermore, we plan to incorporate higher-dimensional sensor modalities, such as vision and touch, for both policy observations and factors of simulation randomization.

\section*{Acknowledgements}
We would like to thank Alexander Lambert, Balakumar Sundaralingam and Giovanni Sutanto for their help with the robot experiments, and  David Ha, James Davidson, Lerrel Pinto and Fabio Ramos for their helpful feedback on the draft of the paper. We would also like to thank the GPU cluster and infrastucture team at NVIDIA for their help all the way through this project.

\bibliographystyle{abbrvunsrtnat} 
\bibliography{simopt}

\begin{thebibliography}{44}
\providecommand{\natexlab}[1]{#1}
\providecommand{\url}[1]{\texttt{#1}}
\expandafter\ifx\csname urlstyle\endcsname\relax
  \providecommand{\doi}[1]{doi: #1}\else
  \providecommand{\doi}{doi: \begingroup \urlstyle{rm}\Url}\fi

\bibitem[Levine et~al.(2018)Levine, Pastor, Krizhevsky, Ibarz, and
  Quillen]{LevinePKIQ18}
S.~Levine, P.~Pastor, A.~Krizhevsky, J.~Ibarz, and D.~Quillen.
\newblock Learning hand-eye coordination for robotic grasping with deep
  learning and large-scale data collection.
\newblock \emph{I. J. Robotics Res.}, 37\penalty0 (4-5):\penalty0 421--436,
  2018.

\bibitem[Pinto and Gupta(2016)]{PintoG16}
L.~Pinto and A.~Gupta.
\newblock Supersizing self-supervision: Learning to grasp from 50k tries and
  700 robot hours.
\newblock In \emph{ICRA}, 2016.

\bibitem[Yahya et~al.(2017)Yahya, Li, Kalakrishnan, Chebotar, and
  Levine]{YahyaLKCL17}
A.~Yahya, A.~Li, M.~Kalakrishnan, Y.~Chebotar, and S.~Levine.
\newblock Collective robot reinforcement learning with distributed asynchronous
  guided policy search.
\newblock In \emph{IROS}, 2017.

\bibitem[Kalashnikov et~al.(2018)Kalashnikov, Irpan, Pastor, Ibarz, Herzog,
  Jang, Quillen, Holly, Kalakrishnan, Vanhoucke, and Levine]{qtopt}
D.~Kalashnikov, A.~Irpan, P.~Pastor, J.~Ibarz, A.~Herzog, E.~Jang, D.~Quillen,
  E.~Holly, M.~Kalakrishnan, V.~Vanhoucke, and S.~Levine.
\newblock Qt-opt: Scalable deep reinforcement learning for vision-based robotic
  manipulation.
\newblock \emph{CoRR}, abs/1806.10293, 2018.

\bibitem[Jakobi et~al.(1995)Jakobi, Husbands, and Harvey]{Jakobi:etal:1995}
N.~Jakobi, P.~Husbands, and I.~Harvey.
\newblock Noise and the reality gap: The use of simulation in evolutionary
  robotics.
\newblock In \emph{European Conference on Artificial Life}. Springer, 1995.

\bibitem[Tobin et~al.(2017)Tobin, Fong, Ray, Schneider, Zaremba, and
  Abbeel]{TobinFRSZA17}
J.~Tobin, R.~Fong, A.~Ray, J.~Schneider, W.~Zaremba, and P.~Abbeel.
\newblock Domain randomization for transferring deep neural networks from
  simulation to the real world.
\newblock In \emph{IROS}, 2017.

\bibitem[Sadeghi and Levine(2017)]{Sadeghi:etal:RSS2017}
F.~Sadeghi and S.~Levine.
\newblock Cad2rl: Real single-image flight without a single real image.
\newblock \emph{RSS}, 2017.

\bibitem[James et~al.(2017)James, Davison, and Johns]{James:etal:CoRL2017}
S.~James, A.~J. Davison, and E.~Johns.
\newblock Transferring end-to-end visuomotor control from simulation to real
  world for a multi-stage task.
\newblock \emph{CoRR}, abs/1707.02267, 2017.

\bibitem[Andrychowicz et~al.(2018)Andrychowicz, Baker, Chociej, Jozefowicz,
  McGrew, Pachocki, Petron, Plappert, Powell, Ray, Schneider, Sidor, Tobin,
  Welinder, Weng, and Zaremba]{handopenai}
M.~Andrychowicz, B.~Baker, M.~Chociej, R.~Jozefowicz, B.~McGrew, J.~Pachocki,
  A.~Petron, M.~Plappert, G.~Powell, A.~Ray, J.~Schneider, S.~Sidor, J.~Tobin,
  P.~Welinder, L.~Weng, and W.~Zaremba.
\newblock Learning dexterous in-hand manipulation.
\newblock \emph{CoRR}, abs/1808.00177, 2018.

\bibitem[{Ljung}(1999)]{Ljung}
L.~{Ljung}.
\newblock \emph{System identification -- theory for the user}.
\newblock Prentice Hall, 1999.

\bibitem[Giri and Bai(2010)]{giri2010block}
F.~Giri and E.-W. Bai.
\newblock \emph{Block-oriented Nonlinear System Identification}.
\newblock London: Springer-Verlag London, 2010.

\bibitem[Deisenroth et~al.(2013)Deisenroth, Neumann, and
  Peters]{Deisenroth2013}
M.~P. Deisenroth, G.~Neumann, and J.~Peters.
\newblock A survey on policy search for robotics.
\newblock \emph{Foundations and Trends in Robotics}, pages 388--403, 2013.

\bibitem[Deisenroth and Rasmussen(2011)]{DeisenrothR11}
M.~P. Deisenroth and C.~E. Rasmussen.
\newblock Pilco: A model-based and data-efficient approach to policy search.
\newblock In \emph{ICML}, 2011.

\bibitem[Deisenroth et~al.(2011)Deisenroth, Rasmussen, and Fox]{DeisenrothRF11}
M.~P. Deisenroth, C.~E. Rasmussen, and D.~Fox.
\newblock Learning to control a low-cost manipulator using data-efficient
  reinforcement learning.
\newblock In \emph{RSS}, 2011.

\bibitem[Kolev and Todorov(2015)]{KolevT15}
S.~Kolev and E.~Todorov.
\newblock Physically consistent state estimation and system identification for
  contacts.
\newblock In \emph{Humanoids}, 2015.

\bibitem[Tzeng et~al.(2015{\natexlab{a}})Tzeng, Devin, Hoffman, Finn, Peng,
  Levine, Saenko, and Darrell]{TzengDHFPLSD15}
E.~Tzeng, C.~Devin, J.~Hoffman, C.~Finn, X.~Peng, S.~Levine, K.~Saenko, and
  T.~Darrell.
\newblock Towards adapting deep visuomotor representations from simulated to
  real environments.
\newblock \emph{CoRR}, abs/1511.07111, 2015{\natexlab{a}}.

\bibitem[Bousmalis et~al.(2017)Bousmalis, Irpan, Wohlhart, Bai, Kelcey,
  Kalakrishnan, Downs, Ibarz, Pastor, Konolige, Levine, and
  Vanhoucke]{graspgan}
K.~Bousmalis, A.~Irpan, P.~Wohlhart, Y.~Bai, M.~Kelcey, M.~Kalakrishnan,
  L.~Downs, J.~Ibarz, P.~Pastor, K.~Konolige, S.~Levine, and V.~Vanhoucke.
\newblock Using simulation and domain adaptation to improve efficiency of deep
  robotic grasping.
\newblock \emph{CoRR}, abs/1709.07857, 2017.

\bibitem[Rusu et~al.(2017)Rusu, Vecerik, Rothörl, Heess, Pascanu, and
  Hadsell]{RusuVRHPH17}
A.~A. Rusu, M.~Vecerik, T.~Rothörl, N.~Heess, R.~Pascanu, and R.~Hadsell.
\newblock Sim-to-real robot learning from pixels with progressive nets.
\newblock In \emph{CoRL}, 2017.

\bibitem[Christiano et~al.(2016)Christiano, Shah, Mordatch, Schneider,
  Blackwell, Tobin, Abbeel, and Zaremba]{ChristianoSMSBT16}
P.~F. Christiano, Z.~Shah, I.~Mordatch, J.~Schneider, T.~Blackwell, J.~Tobin,
  P.~Abbeel, and W.~Zaremba.
\newblock Transfer from simulation to real world through learning deep inverse
  dynamics model.
\newblock \emph{CoRR}, abs/1610.03518, 2016.

\bibitem[Koos et~al.(2010)Koos, Mouret, and Doncieux]{KoosMD10}
S.~Koos, J.-B. Mouret, and S.~Doncieux.
\newblock Crossing the reality gap in evolutionary robotics by promoting
  transferable controllers.
\newblock In \emph{GECCO}. ACM, 2010.

\bibitem[Mordatch et~al.(2015)Mordatch, Lowrey, and Todorov]{MordatchLT15}
I.~Mordatch, K.~Lowrey, and E.~Todorov.
\newblock Ensemble-cio: Full-body dynamic motion planning that transfers to
  physical humanoids.
\newblock In \emph{IROS}, 2015.

\bibitem[Peng et~al.(2018)Peng, Andrychowicz, Zaremba, and
  Abbeel]{PengetalICRA18}
X.~B. Peng, M.~Andrychowicz, W.~Zaremba, and P.~Abbeel.
\newblock Sim-to-real transfer of robotic control with dynamics randomization.
\newblock In \emph{ICRA}, 2018.

\bibitem[Yu et~al.(2017)Yu, Tan, Liu, and Turk]{YuTLT17}
W.~Yu, J.~Tan, C.~K. Liu, and G.~Turk.
\newblock Preparing for the unknown: Learning a universal policy with online
  system identification.
\newblock In \emph{RSS}, 2017.

\bibitem[Muratore et~al.(2018)Muratore, Treede, Gienger, and
  Peters]{MurTreGiePet18}
F.~Muratore, F.~Treede, M.~Gienger, and J.~Peters.
\newblock Domain randomization for simulation-based policy optimization with
  transferability assessment.
\newblock In \emph{CoRL}, 2018.

\bibitem[Wulfmeier et~al.(2017)Wulfmeier, Posner, and Abbeel]{WulfmeierPA17}
M.~Wulfmeier, I.~Posner, and P.~Abbeel.
\newblock Mutual alignment transfer learning.
\newblock \emph{CoRR}, abs/1707.07907, 2017.

\bibitem[Tan et~al.(2018)Tan, Zhang, Coumans, Iscen, Bai, Hafner, Bohez, and
  Vanhoucke]{Tan-RSS-18}
J.~Tan, T.~Zhang, E.~Coumans, A.~Iscen, Y.~Bai, D.~Hafner, S.~Bohez, and
  V.~Vanhoucke.
\newblock Sim-to-real: Learning agile locomotion for quadruped robots.
\newblock In \emph{RSS}, 2018.

\bibitem[Lowrey et~al.(2018)Lowrey, Kolev, Dao, Rajeswaran, and
  Todorov]{LowreyKDRT18}
K.~Lowrey, S.~Kolev, J.~Dao, A.~Rajeswaran, and E.~Todorov.
\newblock Reinforcement learning for non-prehensile manipulation: Transfer from
  simulation to physical system.
\newblock In \emph{SIMPAR}, 2018.

\bibitem[Antonova et~al.(2017)Antonova, Cruciani, Smith, and
  Kragic]{AntonovaCSK17}
R.~Antonova, S.~Cruciani, C.~Smith, and D.~Kragic.
\newblock Reinforcement learning for pivoting task.
\newblock \emph{CoRR}, abs/1703.00472, 2017.

\bibitem[Pinto et~al.(2017)Pinto, Andrychowicz, Welinder, Zaremba, and
  Abbeel]{lerrel}
L.~Pinto, M.~Andrychowicz, P.~Welinder, W.~Zaremba, and P.~Abbeel.
\newblock Asymmetric actor critic for image-based robot learning.
\newblock \emph{CoRR}, abs/1710.06542, 2017.

\bibitem[Tan et~al.(2016)Tan, Xie, Boots, and Liu]{TanXBL16}
J.~Tan, Z.~Xie, B.~Boots, and C.~K. Liu.
\newblock Simulation-based design of dynamic controllers for humanoid
  balancing.
\newblock In \emph{IROS}, 2016.

\bibitem[Zhu et~al.(2018)Zhu, Kimmel, Bekris, and Boularias]{ZhuKBB18}
S.~Zhu, A.~Kimmel, K.~E. Bekris, and A.~Boularias.
\newblock Fast model identification via physics engines for data-efficient
  policy search.
\newblock In \emph{IJCAI}. ijcai.org, 2018.

\bibitem[Farchy et~al.(2013)Farchy, Barrett, MacAlpine, and
  Stone]{AAMAS13-Farchy}
A.~Farchy, S.~Barrett, P.~MacAlpine, and P.~Stone.
\newblock Humanoid robots learning to walk faster: From the real world to
  simulation and back.
\newblock In \emph{AAMAS}, 2013.

\bibitem[Hanna and Stone(2017)]{AAAI17-Hanna}
J.~Hanna and P.~Stone.
\newblock Grounded action transformation for robot learning in simulation.
\newblock In \emph{AAAI}, 2017.

\bibitem[Rajeswaran et~al.(2016)Rajeswaran, Ghotra, Levine, and
  Ravindran]{RajeswaranGLR16}
A.~Rajeswaran, S.~Ghotra, S.~Levine, and B.~Ravindran.
\newblock Epopt: Learning robust neural network policies using model ensembles.
\newblock \emph{CoRR}, abs/1610.01283, 2016.

\bibitem[Kingma and Welling(2013)]{KingmaW13}
D.~P. Kingma and M.~Welling.
\newblock Auto-encoding variational {B}ayes.
\newblock \emph{CoRR}, abs/1312.6114, 2013.

\bibitem[Schulman et~al.(2017)Schulman, Wolski, Dhariwal, Radford, and
  Klimov]{SchulmanWDRK17}
J.~Schulman, F.~Wolski, P.~Dhariwal, A.~Radford, and O.~Klimov.
\newblock Proximal policy optimization algorithms.
\newblock \emph{CoRR}, abs/1707.06347, 2017.

\bibitem[Liang et~al.(2018)Liang, Makoviychuk, Handa, Chentanez, Macklin, and
  Fox]{Liang:etal:CoRL2018}
J.~Liang, V.~Makoviychuk, A.~Handa, N.~Chentanez, M.~Macklin, and D.~Fox.
\newblock Gpu-accelerated robotic simulation for distributed reinforcement
  learning.
\newblock \emph{CoRL}, 2018.

\bibitem[Peters et~al.(2010)Peters, Mülling, and Altun]{PetersMA10}
J.~Peters, K.~Mülling, and Y.~Altun.
\newblock Relative entropy policy search.
\newblock In \emph{AAAI}, 2010.

\bibitem[Tzeng et~al.(2015{\natexlab{b}})Tzeng, Hoffman, Darrell, and
  Saenko]{TzengHDS15}
E.~Tzeng, J.~Hoffman, T.~Darrell, and K.~Saenko.
\newblock Simultaneous deep transfer across domains and tasks.
\newblock In \emph{ICCV}, 2015{\natexlab{b}}.

\bibitem[Sermanet et~al.(2018)Sermanet, Lynch, Chebotar, Hsu, Jang, Schaal, and
  Levine]{tcn}
P.~Sermanet, C.~Lynch, Y.~Chebotar, J.~Hsu, E.~Jang, S.~Schaal, and S.~Levine.
\newblock Time-contrastive networks: Self-supervised learning from video.
\newblock In \emph{ICRA}, 2018.

\bibitem[Goodfellow et~al.(2014)Goodfellow, Pouget-Abadie, Mirza, Xu,
  Warde-Farley, Ozair, Courville, and Bengio]{GoodfellowPMXWOCB14}
I.~J. Goodfellow, J.~Pouget-Abadie, M.~Mirza, B.~Xu, D.~Warde-Farley, S.~Ozair,
  A.~C. Courville, and Y.~Bengio.
\newblock Generative adversarial nets.
\newblock In \emph{NIPS}, 2014.

\bibitem[Ho and Ermon(2016)]{HoE16}
J.~Ho and S.~Ermon.
\newblock Generative adversarial imitation learning.
\newblock In \emph{NIPS}, 2016.

\bibitem[Hausman et~al.(2017)Hausman, Chebotar, Schaal, Sukhatme, and
  Lim]{HausmanCSSL17}
K.~Hausman, Y.~Chebotar, S.~Schaal, G.~S. Sukhatme, and J.~J. Lim.
\newblock Multi-modal imitation learning from unstructured demonstrations using
  generative adversarial nets.
\newblock In \emph{NIPS}, 2017.

\bibitem[Schmidt et~al.(2014)Schmidt, Newcombe, and Fox]{Schmidt:etal:RSS2014}
T.~Schmidt, R.~A. Newcombe, and D.~Fox.
\newblock Dart: Dense articulated real-time tracking.
\newblock In \emph{RSS}, 2014.

\end{thebibliography}
\clearpage

\appendix

\subsection{Comparison to trajectory-based parameter learning}
In our work, we run a closed-loop policy in simulation to obtain simulated roll-outs for \textit{SimOpt} optimization. Alternatively, we could directly set the simulator to states and execute actions from the real world trajectories as proposed in~\cite{TanXBL16,ZhuKBB18}. However, such a setting is not always possible as we might not be able to observe all required variables for setting the internal state of the simulator at each time point, \textit{e.g.} the current bending configuration of the rope in the swing-peg-in-hole task, which we are able to initialize but can not continually track with our real world set up. 

Without being able to set the simulator to the real world states continuously, we still can try to copy the real world actions and execute them in an open-loop manner in simulation. However, in our simulated experiments we notice that especially when making particular state dimensions unobservable for \textit{SimOpt} cost computation, such as X-position of the cabinet in the drawer opening task, executing a closed-loop policy still leads to meaningful simulation parameter updates compared to the open-loop execution. We believe in this case the robot behavior is still dependent on the particular simulated scenario due to the closed-loop nature of the policy, which also reflects in the joint trajectories of the robot that are still included in the \textit{SimOpt} cost function. This means that by using a closed-loop policy we can still update the simulation parameter distribution even without explicitly including some of the relevant observations in the \textit{SimOpt} cost computation.

\subsection{Simulation parameters}
\label{app:sim_params}
Tables \ref{tab:drawer_params} and \ref{tab:peg_params} show the initial mean, diagonal values of the initial covariance matrix and the final mean of the Gaussian simulation parameter distributions that have been optimized with \textit{SimOpt} in drawer opening (Table \ref{tab:drawer_params}) and swing-peg-in-hole (Table \ref{tab:peg_params}) tasks. 
\begin{table}[htbp]
\footnotesize
\begin{center}
\bgroup
\def\arraystretch{1.3}
\setlength{\tabcolsep}{2.5pt}
\begin{tabular}{|l|c|c||c|}
\hline
& $\mu_{init}$ & $\text{diag}(\Sigma_{init})$ & $\mu_{final}$\\ \hline
\multicolumn{4}{|l|}{\textbf{Robot properties}} \\ \hline
Joint compliance (7D) & $[\text{-}6.0\dotsc\text{-}6.0]$ & 0.5 & $[\text{-}6.5\dotsc\text{-}6.1]$ \\ \hline
Joint damping (7D)& $[3.0\dotsc3.0]$ & 0.5 & $[2.4\dotsc2.7]$ \\ \hline
Gripper compliance & -11.0 & 0.5 & -10.9\\ \hline
Gripper damping & 0.0 & 0.5 & 0.34\\ \hline
Joint action scaling (7D)&  $[0.26\dotsc0.26]$ & 0.01 &  $[0.19\dotsc0.35]$  \\ \hline
\multicolumn{4}{|l|}{\textbf{Cabinet properties}}\\ \hline
 Drawer joint compliance & 7.0&1.0 & 8.3  \\ \hline
 Drawer joint damping &  2.0&0.5 & 0.81 \\ \hline
Drawer handle friction & 0.001 & 0.5 & 2.13 \\ \hline
\end{tabular}
\egroup
\end{center}
\caption{Drawer opening: simulation parameter distribution.}
\label{tab:drawer_params}

\end{table}

\begin{table}[htbp]
\footnotesize
\begin{center}
\bgroup
\def\arraystretch{1.3}
\setlength{\tabcolsep}{2.7pt}
\begin{tabular}{|l|c|c||c|}
\hline
& $\mu_{init}$ & $\text{diag}(\Sigma_{init})$ & $\mu_{final}$\\ \hline
\multicolumn{4}{|l|}{\textbf{Robot properties}} \\ \hline
Joint compliance (7D) & $[\text{-}8.0\dotsc\text{-}8.0]$ & 1.0 & $[\text{-} 8.2\dotsc\text{-}7.8]$ \\ \hline
Joint damping (7D) & $[\text{-}3.0\dotsc\text{-}3.0]$  & 1.0 &  $[\text{-}3.0\dotsc\text{-}2.6]$\\ \hline
Joint action scaling (7D) & $[0.5\dotsc0.5]$ & 0.02 & $[0.25\dotsc0.44]$\\ \hline
\multicolumn{4}{|l|}{\textbf{Rope properties}}\\ \hline
Rope torsion compliance & 2.0 & 0.07 & 1.89\\ \hline
Rope torsion damping & 0.1 & 0.07 & 0.48\\ \hline
Rope bending compliance & 10.0 & 0.5 & 9.97\\ \hline
Rope bending damping & 0.01 & 0.05 & 0.49\\ \hline
Rope segment width & 0.004 & $2\mathrm{e}\text{-}4$ & 0.007\\ \hline
Rope segment length & 0.016 & 0.004  & 0.017\\ \hline
Rope segment friction & 0.25 & 0.03 & 0.29\\ \hline
Rope density & 2500.0 & 8.0 & 2500.12\\ \hline
\multicolumn{4}{|l|}{\textbf{Peg properties}}\\ \hline
Peg scale & $0.33$ & 0.01 & $0.30$\\ \hline
Peg friction & 1.0 & 0.06 & 1.0 \\ \hline
Peg mass coefficient & 1.0 & 0.06 & 1.06\\ \hline
Peg density & 400.0 & 10.0 & 400.07\\ \hline
\multicolumn{4}{|l|}{\textbf{Peg box properties}}\\ \hline
Peg box scale & 0.029 & 0.01 & 0.034\\ \hline
Peg box friction & 1.0 & 0.2  & 1.01\\ \hline
\end{tabular}
\egroup
\end{center}
\caption{Swing-peg-in-hole: simulation parameter distribution.}
\label{tab:peg_params}
\vspace*{4.5in}
\end{table}

\newpage
\subsection{SimOpt parameters}
Tables \ref{tab:app_peg} and \ref{tab:app_cabinet} show the \textit{SimOpt} distribution update parameters for swing-peg-in-hole and drawer opening tasks including REPS~\cite{PetersMA10} parameters, settings of the discrepancy function $D(\tau^{ob}_\xi, \tau^{ob}_{real})$, weights of each observation dimension in the discrepancy function, and reinforcement learning settings such as parallelized PPO~\cite{SchulmanWDRK17,Liang:etal:CoRL2018} training parameters and task reward weights.

\begin{table}[h]
\footnotesize
\begin{center}
\bgroup
\def\arraystretch{1.3}
\setlength{\tabcolsep}{5pt}
\begin{tabular}{|l|c|}
\hline 
\multicolumn{2}{|l|}{\textbf{Simulation distribution update parameters}} \\ \hline
Number of REPS updates per SimOpt iteration & 3 \\ \hline
Number of simulation parameter samples per update & 9600 \\ \hline
Timesteps per simulation parameter sample & 453 \\ \hline
KL-threshold & 1.0 \\ \hline
Minimum temperature of sample weights & 0.001 \\ \hline
\multicolumn{2}{|l|}{\textbf{Discrepancy function parameters}} \\ \hline
L1-cost weight & 0.5 \\ \hline
L2-cost weight & 1.0 \\ \hline
Gaussian smoothing standard deviation (timesteps) & 5 \\ \hline
Gaussian smoothing truncation (timesteps) & 4 \\ \hline
\multicolumn{2}{|l|}{\textbf{Observation dimensions cost weights}} \\ \hline
Joint angles (7D) & 0.05 \\ \hline
Peg position (3D) & 1.0 \\ \hline
Peg position in the previous timestep (3D) & 1.0 \\ \hline
\multicolumn{2}{|l|}{\textbf{PPO parameters}} \\ \hline
Number of agents & 100 \\ \hline
Episode length & 150 \\ \hline
Timesteps per batch & 64 \\ \hline
Clip parameter & 0.2 \\ \hline
$\gamma$ & 0.99 \\ \hline
$\lambda$ & 0.95 \\ \hline
Entropy coefficient & 0.0 \\ \hline
Optimization epochs & 10 \\ \hline
Optimization batch size per agent & 8\\ \hline
Optimization step size &  $5\mathrm{e}\text{-}4$ \\ \hline
Desired KL-step & 0.01 \\ \hline
\multicolumn{2}{|l|}{\textbf{RL reward weights}} \\ \hline
L1-distance between the peg and the hole & -10.0 \\ \hline
L2-distance between the peg and the hole & -4.0 \\ \hline
Task solved (peg completely in the hole) bonus & 0.1 \\ \hline
Action penalty & -0.7 \\ \hline
\end{tabular}
\egroup
\end{center}
\caption{Swing-peg-in-hole: SimOpt parameters.}
\label{tab:app_peg}
\end{table}

\begin{table}[htbp]
\footnotesize
\begin{center}
\bgroup
\def\arraystretch{1.3}
\setlength{\tabcolsep}{2.6pt}
\begin{tabular}{|l|c|}
\hline 
\multicolumn{2}{|l|}{\textbf{Simulation distribution update parameters}} \\ \hline
Number of REPS updates per SimOpt iteration & 20 \\ \hline
Number of simulation parameter samples per update & 9600 \\ \hline
Timesteps per simulation parameter sample & 453 \\ \hline
KL-threshold & 1.0 \\ \hline
Minimum temperature of sample weights & 0.001 \\ \hline
\multicolumn{2}{|l|}{\textbf{Discrepancy function parameters}} \\ \hline
L1-cost weight & 0.5 \\ \hline
L2-cost weight & 1.0 \\ \hline
Gaussian smoothing standard deviation (timesteps) & 5 \\ \hline
Gaussian smoothing truncation (timesteps) & 4 \\ \hline
\multicolumn{2}{|l|}{\textbf{Observation dimensions cost weights}} \\ \hline
Joint angles (7D) & 0.5 \\ \hline
Drawer position (3D) & 1.0 \\ \hline
\multicolumn{2}{|l|}{\textbf{PPO parameters}} \\ \hline
Number of agents & 400 \\ \hline
Episode length & 150 \\ \hline
Timesteps per batch & 151 \\ \hline
Clip parameter & 0.2 \\ \hline
$\gamma$ & 0.99 \\ \hline
$\lambda$ & 0.95 \\ \hline
Entropy coefficient & 0.0 \\ \hline
Optimization epochs & 5 \\ \hline
Optimization batch size per agent & 8\\ \hline
Optimization step size &  $5\mathrm{e}\text{-}4$ \\ \hline
Desired KL-step & 0.01 \\ \hline
\multicolumn{2}{|l|}{\textbf{RL reward weights}} \\ \hline
L2-distance between end-effector and drawer handle & -0.5 \\ \hline
Angular alignment of end-effector with drawer handle & -0.07 \\ \hline
Opening distance of the drawer & -0.4 \\ \hline
Keeping fingers around the drawer handle bonus & 0.005 \\ \hline
Action penalty & -0.005 \\ \hline
\end{tabular}
\egroup
\end{center}
\caption{Drawer opening: SimOpt parameters.}
\label{tab:app_cabinet}
\vspace*{3.2in}
\end{table}

\end{document}